\documentclass[letterpaper]{article} 
\usepackage{aaai24}  
\usepackage{times}  
\usepackage{helvet}  
\usepackage{courier}  
\usepackage[hyphens]{url}  
\usepackage{graphicx} 
\usepackage{natbib}  
\usepackage{caption} 
\usepackage{algorithm}
\usepackage{algorithmic}
\usepackage{multirow}
\usepackage{amsfonts}

\usepackage{amsmath}
\usepackage{amssymb}
\usepackage{booktabs}
\usepackage{adjustbox}
\usepackage{makecell}
\usepackage{bbding}
\usepackage{pifont}
\usepackage{wasysym}
\usepackage{utfsym}
\usepackage{fontawesome}

\usepackage{caption}
\usepackage{xcolor}
\definecolor{citecolor}{HTML}{2980b9}
\definecolor{linkcolor}{HTML}{c0392b}

\usepackage{colortbl}
\usepackage{enumerate}
\usepackage{float}
\usepackage{subfig}

\makeatletter
  \newcommand\figcaption{\def\@captype{figure}\caption}
  \newcommand\tabcaption{\def\@captype{table}\caption}
\makeatother

\usepackage[capitalize]{cleveref}
\crefname{section}{Sec.}{Secs.}
\Crefname{section}{Section}{Sections}
\Crefname{table}{Table}{Tables}
\crefname{table}{Tab.}{Tabs.}

\usepackage{newfloat}
\usepackage{listings}
\DeclareCaptionStyle{ruled}{labelfont=normalfont,labelsep=colon,strut=off} 
\floatstyle{ruled}
\newfloat{listing}{tb}{lst}{}
\floatname{listing}{Listing}
\nocopyright 

\title{Less is More: Towards Efficient Few-shot 3D Semantic Segmentation via Training-free Networks}
\author{
Xiangyang Zhu$^{*1}$, Renrui Zhang$^{*\dagger\ddagger2,4}$, Bowei He$^{1}$, Ziyu Guo$^{2}$, Jiaming Liu$^{3}$, \\ Hao Dong$^{3}$, Peng Gao$^{4}$ \vspace{0.2cm}
}
\affiliations{
    $^1$City University of Hong Kong \quad \vspace{0.07cm}
  $^2$The Chinese University of Hong Kong\\
  $^3$Peking University \quad \vspace{0.2cm} $^4$Shanghai Artificial Intelligence Laboratory\\

    \normalsize{$^*$ Equal contribution}\quad  $\dagger$ Project leader\quad  $\ddagger$ Corresponding author\vspace{0.2cm}\\

    \texttt{\{xiangyzhu6-c, boweihe2-c\}@my.cityu.edu.hk},\\
\texttt{\{zhangrenrui, gaopeng\}@pjlab.org.cn} \vspace{0.3cm}
}

\usepackage{bibentry}

\begin{document}

\maketitle

\begin{abstract}
To reduce the reliance on large-scale datasets, recent works in 3D segmentation resort to few-shot learning. Current 3D few-shot semantic segmentation methods first pre-train the models on `seen' classes, and then evaluate their generalization performance on `unseen' classes. However, the prior pre-training stage not only introduces excessive time overhead, but also incurs a significant domain gap on `unseen' classes. To tackle these issues, we propose an efficient \textbf{T}raining-free \textbf{F}ew-shot 3D \textbf{S}egmentation netwrok, \textbf{TFS3D}, and a further training-based variant, \textbf{TFS3D-T}. Without any learnable parameters, TFS3D extracts dense representations by trigonometric positional encodings, and achieves comparable performance to previous training-based methods. Due to the elimination of pre-training, TFS3D can alleviate the domain gap issue and save a substantial amount of time. Building upon TFS3D, TFS3D-T only requires to train a lightweight query-support transferring attention (QUEST), which enhances the interaction between the few-shot query and support data. Experiments demonstrate TFS3D-T improves previous state-of-the-art methods by \textbf{+6.93\%} and \textbf{+17.96\%} mIoU respectively on S3DIS and ScanNet, while reducing the training time by  \textbf{-90\%}, indicating superior effectiveness and efficiency. Code is available at \textcolor[rgb]{1,0.41,0.71}{\url{https://github.com/yangyangyang127/TFS3D}}.
\end{abstract}

\vspace{-0.1cm}
\section{Introduction}

Point cloud segmentation is an essential procedure in autonomous driving~\cite{krispel2020fuseseg, wang2020pillar}, robotics~\cite{lewandowski2019fast, ahmed2018edge}, and other computer vision or graphic applications~\cite{yue2018lidar, bello2020deep}. Recently, to achieve favorable point-level segmentation, many learning-based methods have been proposed and achieved satisfactory results on various 3D benchmarks~\cite{li2022primitive3d, lai2022stratified, yang2023swin3d,  li2023transformer}. However, such algorithms require the construction of large-scale and accurately annotated datasets, which is expensive and time-consuming. 

\begin{figure}[t]
\subfloat[Existing Methods]{\includegraphics[width=0.47\textwidth]{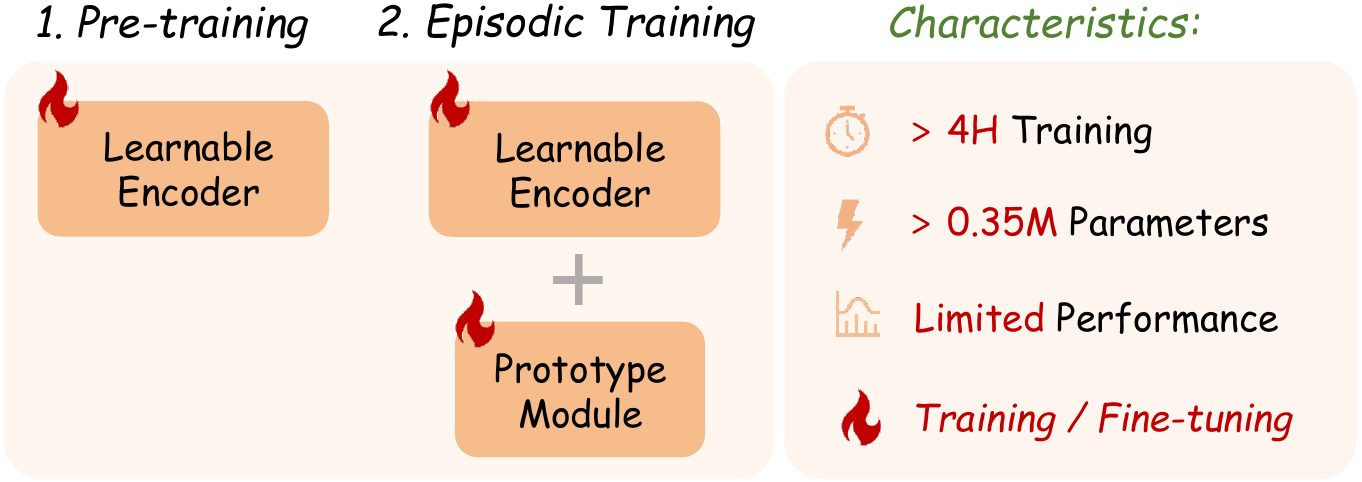}} \\
\subfloat[Our \textbf{TFS3D} or \textbf{TFS3D-T}]{\includegraphics[width=0.47\textwidth]{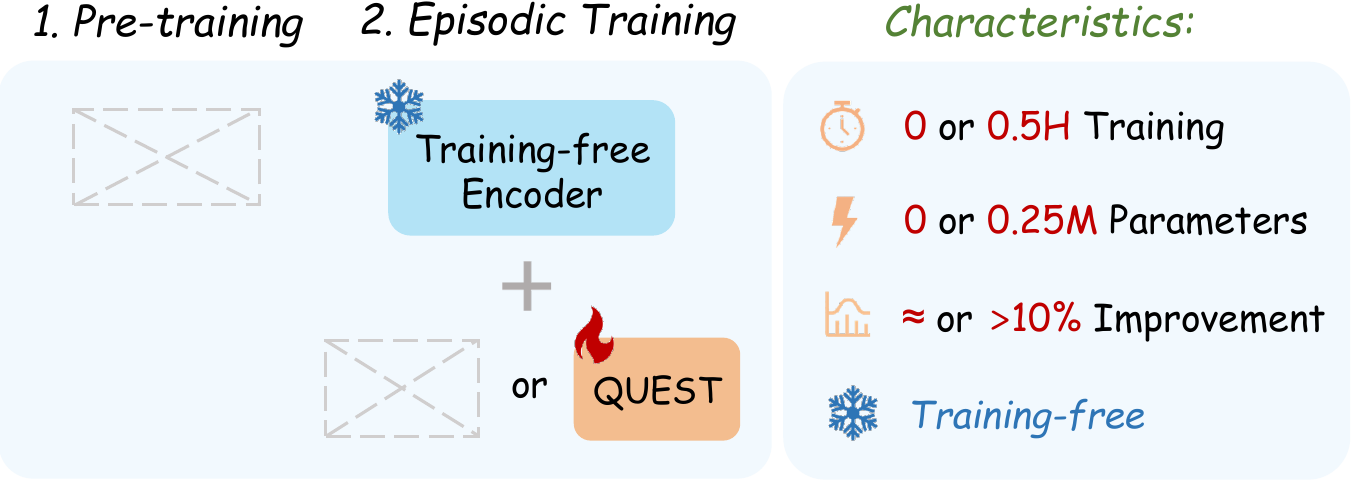}}
\vspace{-0.1cm}
\caption{\textbf{Comparison of Existing Methods and Our Approaches.} Our training-free TFS3D contains no learnable parameters and thus discards both pre-training and episodic training stages with superior efficiency. Our training-based TFS3D-T further improves the performance by a lightweight QUEST attention module.}
\label{fig:existing_comparison}
\vspace{-0.3cm}
\end{figure}

The data-hungry problem can be effectively mitigated by the few-shot learning strategy, which has garnered significant attention in the 3D community~\cite{he2023prototype}. 
As shown in Fig. \ref{fig:existing_comparison} (a), existing 3D few-shot semantic segmentation methods basically follow the meta-learning scheme to learn a 3D encoder and a prototype generation module. They mainly take three steps to achieve the final goal:  
\begin{enumerate}
  \item \textbf{Pre-training on `seen' classes} by supervised learning. Considering the lack of pre-trained models in the 3D field, this step trains a learnable 3D encoder, e.g., DGCNN, to obtain the ability for extracting general 3D point cloud representations.
  \item \textbf{Episodic training on `seen' classes} to fit the query-support few-shot segmentation tasks. In this step, the pre-trained encoder is appended with a learnable prototype module and fine-tuned to extract discriminative prototypes from the support set, which are utilized to guide the semantic segmentation of query sets.
  \item \textbf{Testing on `unseen' classes} to evaluate the model. After episodic training, the model is evaluated on test episodes that contain unseen classes. The model is expected to segment new classes by the same query-support paradigms as the episodic training.
\end{enumerate}
However, this pipeline encompasses two noteworthy issues: 1) the learnable encoder being pre-trained and fine-tuned on `seen' classes will inevitably introduce a remarkable domain gap when evaluated on `unseen' classes; 2) the complexity of the training process, including pre-training and episodic training, incurs substantial time and resource overhead.

\begin{figure}[t]
\centering
\subfloat[Performance Difference of Seen and Unseen Classes]{\includegraphics[width=0.22\textwidth]{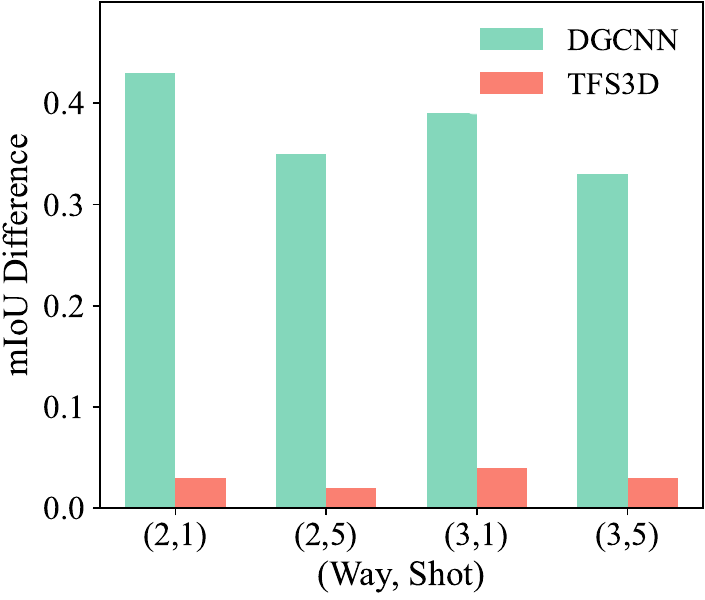}} 
\hspace{4pt}
\subfloat[KL Divergence between the Support Set and Query Set]{\includegraphics[width=0.22\textwidth]{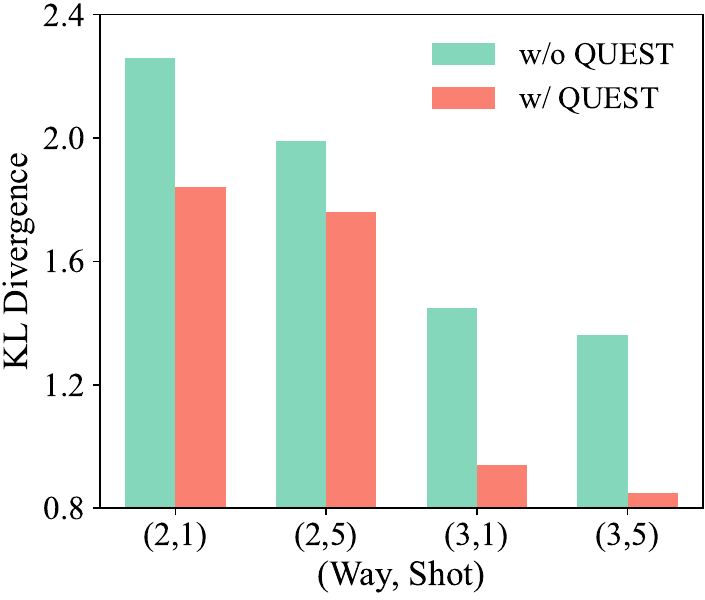}}
\vspace{-0.15cm}
\caption{\textbf{Efficacy of Alleviating Domain Gap by TFS3D (a) and Prototype Bias by TFS3D-T (b)} on S3DIS~\cite{armeni20163d} dataset.
The horizontal axis presents the number of ways and shots in the form of (Way, Shot). 
}
\label{fig:efficacy_QUEST}
\vspace{-0.4cm}
\end{figure}

To address these issues, we propose a \textbf{T}raining-free \textbf{F}ew-shot \textbf{S}egmentation framework, \textbf{TFS3D}, which is both efficient and effective.
Inspired by Point-NN~\cite{zhang2023parameter}, TFS3D utilizes a non-parametric and training-free encoder for feature encoding, which stacks trigonometric positional encodings to project raw point coordinates with RGB signals into embedding space. Then, we integrate the few-shot support-set features to produce category prototypes, and leverage them to predict the segmentation masks for the query set by similarity matching. As shown in Fig. \ref{fig:existing_comparison} (a), TFS3D purely contains no parameter and discards all two stages of pre-training and episodic training, and performs even comparably to some existing training-based methods.
Such a training-free property significantly simplifies the few-shot training pipeline with minimal resource consumption, and mitigates the domain gap caused by different training-test categories. As visualized in Fig. \ref{fig:efficacy_QUEST} (a), we observe that TFS3D shows marginal performance difference between seen and unseen categories, while the widely adopted DGCNN~\cite{wang2019dynamic} encoder presents a much worse generalization ability due to cross-domain training and testing.


On top of this, we also propose a \textbf{T}raining-based variant of our approach, \textbf{TFS3D-T}, to further boost the performance by efficient training. In detail, we inherit the training-free encoder of TFS3D and append an additional trainable \textbf{QUE}ry-\textbf{S}upport \textbf{T}ransferring attention module, termed as \textbf{QUEST}. QUEST enhances the prototypes in the support-set domain by the knowledge of the query-set domain, which suppresses the prototype bias issue caused by the small few-shot support set. As shown in Fig. \ref{fig:efficacy_QUEST} (b), the reduced query-support distribution differences suggest that the prototypes are shifted to the query-set domain.
TFS3D-T only learns the QUEST attention module and does not require pre-training just as the training-free TFS3D, as shown in Fig. \ref{fig:existing_comparison} (b). Experiments show that TFS3D-T achieves new state-of-the-art (SOTA) performance on both S3DIS~\cite{armeni20163d} and ScanNet~\cite{dai2017scannet} datasets, surpassing the second-best by \textbf{+6.93\%} and \textbf{+17.96\%}, respectively, while reducing the training time by over \textbf{-90\%}.

In summary, our contributions are as follows:

\begin{itemize}
    \item We introduce a training-free few-shot learning framework, TFS3D, for 3D point cloud semantic segmentation, which can also serve as a basis to construct the training-based better-performed variant, TFS3D-T.
    \item We design a new query-support interaction module, QUEST, in TFS3D-T to adapt category prototypes by learning the affinity between support and query features.
    \item  Comprehensive experiments are conducted to verify the efficiency and efficacy of our proposed method. We achieve state-of-the-art results with the least parameters and a substantially simplified learning pipeline.
\end{itemize}

\section{Related Work}

\paragraph{Point Cloud Semantic Segmentation}aims to assign each point with the correct category label within a pre-defined label space. The early PointNet and PointNet++ establish the basic framework for learning-based 3D analysis~\cite{qi2017pointnet, qi2017pointnet++}. The follow-up works~\cite{zhang2023parameter,zhao2021point,li2023transformer,zhang2022point,guo2023joint} further improve the performance of segmentation and other 3D tasks~\cite{misra20213detr,chi2023bev,yu2021pointr}. Despite this, these methods are data-hungry, and require extensive additional labeled data to be fine-tuned for unseen classes. To address this issue, a series of 3D few-shot learning methods have been proposed~\cite{mao2022bidirectional, zhang2023few}. AttMPTI~\cite{zhao2021few} extracts multiple prototypes from support-set features and transductively predicts query labels. 2CBR~\cite{zhu2023cross} proposes cross-class rectification to alleviate the support-query domain gap, and PAP-FZ3D~\cite{he2023prototype} jointly trains few-shot and zero-shot semantic segmentation tasks. All of these methods adopt the conventional meta-learning strategy including both pre-training and episodic training stages. In this work, we propose more efficient solutions for 3D few-shot semantic segmentation. We first devise a training-free encoder without parameters, thus discarding the time-consuming pre-training. 
Then, we propose a training-free framework, TFS3D, and a training-based variant, TFS3D-T, achieving competitive performance with minimal resources and simplifying the traditional meta-learning pipelines.

\paragraph{Positional Encoding}(PE) projects a location vector into a high dimensional embedding that can preserve spatial information and, at the same time, be learning-friendly for downstream algorithms \cite{mai2022review}. Transformer~\cite{vaswani2017attention} first utilize PE to indicate the one-dimensional location of parallel input entries in a sequence, which is composed of trigonometric functions. Such trigonometric PE can encode both absolute and relative positions, and each of its dimensions corresponds to a predefined frequency and phase, which has also been employed for learning high-frequency functions~\cite{tancik2020fourier} and improving 3D rendering in NeRF~\cite{mildenhall2021nerf}. 
In 3D domains, Point-NN~\cite{zhang2023parameter,zhang2022nearest} leverages the basic trigonometric PE to encode point coordinates for robust shape classification.
Inspired by them, we utilize trigonometric PE to encode both positional and color information of point clouds for more comprehensive 3D feature extraction. We also utilize the advantages of different PE frequency distributions for more discriminative geometry encoding.

\begin{figure*}[t]
\centering
\includegraphics[width=0.98\textwidth]{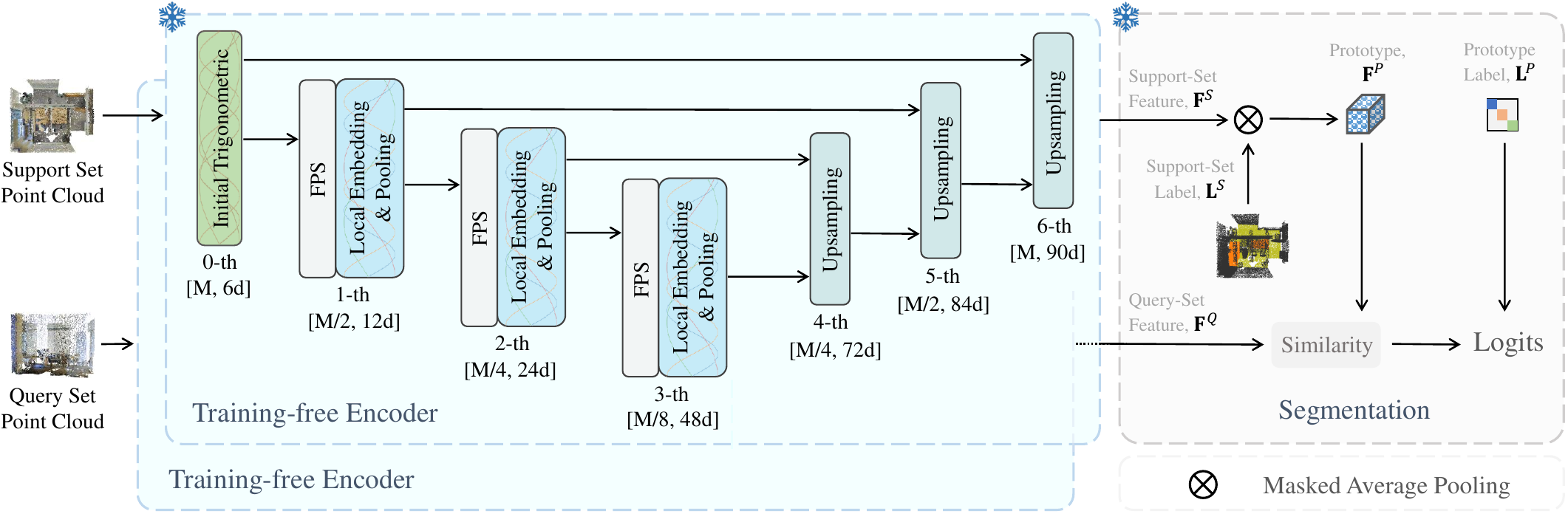}
\caption{\textbf{Framework of the Training-free TFS3D}, including a training-free encoder and a segmentation head. The encoder extracts the support- and query-set features and the segmentation head segment the query set based on similarity matching.}
\label{fig:framework}
\vspace{-0.3cm}
\end{figure*}

\section{Problem Definition}

 

\label{sec:problem_definition}

We first illustrate the task definition of few-shot 3D semantic segmentation. 
We follow previous works \cite{zhao2021few, he2023prototype} to adopt the popular episodic training/testing paradigm \cite{vinyals2016matching} after the pre-training stage. Each episode is instantiated as an $N$-way $K$-shot task, which contains a support set and a query set. The support set comprises $N$ target classes, and each class corresponds to $K$ point cloud samples with their point-level segmentation labels. The query set contains a fixed number of point clouds that need to be segmented.
Each episodic task aims to segment the query-set point clouds into $N$ target classes along with a ``background'' class based on the guidance of the support set.

To achieve this, we regard the semantic segmentation task as a point-level similarity-matching problem. We first utilize a feature encoder to extract the features of support-set point cloud samples, and generate $N+1$ prototypes for all $N+1$ classes. Then, we adopt the same feature encoder to obtain the feature of every point in query samples, and conduct similarity matching with the prototypes for point-level classification. In this way, the query-set point clouds can be segmented into $N+1$ semantic categories.

In the following two sections, we respectively illustrate the details of our proposed efficient frameworks, the training-free TFS3D and training-based TFS3D-T.



\section{Training-free TFS3D}

The detailed structure of TFS3D is shown in Fig. \ref{fig:framework}, which contains a U-Net style training-free encoder and a similarity-based segmentation head. The encoder embeds the point clouds into high-dimensional representations, and the segmentation head conducts non-parametric similarity matching to output the final prediction. The entire framework does not introduce any learnable parameters, and thus discards both two stages of pre-training and episodic training, differing from existing algorithms using DGCNN~\cite{wang2019dynamic} as the learnable encoder. 
Following, we respectively describe the details of the encoder and segmentation head. 


\subsection{TFS3D Encoder}

Given a point cloud $\left \{ \mathbf{p}_{i} \right \}_{i=1}^{M}$ containing $M$ points, our goal is to encode each point into embedding space. Taking a random point $\mathbf{p}=(x,y,z)$ as an example, we denote its RGB color as $\mathbf{c}=(r, g, b)$. The encoder aims to extract shape knowledge based on both coordinate and color information. As shown in Fig. \ref{fig:framework}, motivated by Point-NN~\cite{zhang2023parameter}, it comprises three types of layers: \textbf{\textit{Initial Trigonometric}} projects $\mathbf{p}$ into embedding space; stacked \textbf{\textit{Local Embedding}} layers extract hierarchical 3D features by trigonometric positional encoding (PE); \textbf{\textit{Upsampling}} layers upsample hierarchical features into the input point number for obtaining the final point-level features.

\paragraph{\textit{Initial Trigonometric.}} At the start of the encoder, we project the coordinate $\mathbf{p}$ and color $\mathbf{c}$ into high-dimensional space via trigonometric form PEs with $d$ frequencies $\mathbf{u}=[u_1, ..., u_d]$, denoted as $\operatorname{Emb}(\cdot)$, then
\begin{equation}
\setlength{\abovedisplayskip}{4pt}
\setlength{\belowdisplayskip}{3pt}
\begin{aligned}
\operatorname{Emb}(\mathbf{p};\mathbf{u}) = [ & \sin(2 \pi \mathbf{u} \mathbf{p}), \cos(2 \pi \mathbf{u} \mathbf{p})]  \quad \in \mathbb{R}^{6d}, \\
\operatorname{Emb}(\mathbf{c};\mathbf{u}) = [ & \sin(2 \pi \mathbf{u} \mathbf{c}),  \cos(2 \pi \mathbf{u} \mathbf{c})] \quad \in \mathbb{R}^{6d},
\end{aligned}
\label{equ:sine_embed}
\end{equation}
where $6d$ comes from the combination of 3 coordinates, $(x,y,z)$, with 2 functions, $\sin(\cdot)$ and $\cos(\cdot)$. All frequency components in $\mathbf{u}$ adhere to log-linear form, $u_{i}=\theta^{i/d}, i=1,..., d$, where the scale $\theta$ and dimension $d$ are two hyperparameters.
These log-linear spaced frequencies cover a wide band and enable efficient capture of both coarse- and fine-grained geometrics.
We designate this step as the 0-th layer and utilize $\mathbf{f}^{0,p} = \operatorname{Emb}(\mathbf{p};\mathbf{u})$ and $\mathbf{f}^{0,c} = \operatorname{Emb}(\mathbf{c};\mathbf{u})$ to denote the coordinate and color vectors, respectively.
We then merge $\mathbf{f}^{0,p}$ and $\mathbf{f}^{0,c}$ to obtain a comprehensive initial embedding via $\alpha$-weighted summation,
\begin{equation}
\setlength{\abovedisplayskip}{5pt}
\setlength{\belowdisplayskip}{5pt}
\begin{aligned}
\mathbf{f}^{0} = \alpha \cdot \mathbf{f}^{0,p} + (1-\alpha) \cdot \mathbf{f}^{0,c} \quad \in \mathbb{R}^{6d}.
\end{aligned}
\label{equ:initial_integrate}
\end{equation}
These trigonometric initial embeddings can effectively represent coordinate and color information. 

\paragraph{\textit{Local Embedding.}} As presented in Fig. \ref{fig:framework}, our training-free encoder consists of three local embedding layers. Before each layer, we downsample the point cloud by half to increase the receptive field with the Farthest Point Sampling (FPS). The three local embedding layers encode local patterns and output hierarchical features for the point cloud. 
In detail, we also denote $\mathbf{p}$ as a local center sampled by FPS, and consider its neighborhood $\mathcal{N}_p$ searched by the $k$-Nearest Neighbor ($k$-NN) algorithm. We first concatenate the neighbor point feature with the center point feature along the channel dimension,
\begin{equation}
\setlength{\abovedisplayskip}{4pt}
\setlength{\belowdisplayskip}{3pt}
\begin{aligned}
\hat{\mathbf{f}}_{j}^{l} = \operatorname{Concat}(\mathbf{f}^{l-1}, \mathbf{f}_{j}^{l-1}), \quad j \in \mathcal{N}_p,
\end{aligned}
\label{equ:local_concat}
\end{equation}
where the subscript $j$ represents the $j$-th point in $\mathbf{p}$'s neighborhood $\mathcal{N}_p$, and $l \in \left \{ 1, 2, 3 \right \}$ denotes the $l$-th layer as in Fig. \ref{fig:framework}. After concatenation, the expanded embedding $\hat{\mathbf{f}}_{j}^{l} \in \mathbb{R}^{2^{l} \times 6d}$ can incorporate both center and neighbor information. The dimensionality is increased as $2^{l}\times 6d$, since we conduct Eq. \ref{equ:local_concat} at every local embedding layer, each of which doubles the channel dimension.
Then, to reveal local patterns, we refer to Point-NN~\cite{zhang2023parameter} and aggregate the relative coordinate and color information. The normalized relative coordinate $\Delta \mathbf{p}_{j}$ from neighbor point $j$ to the center $\mathbf{p}$ is embedded via PEs as in Eq. \ref{equ:sine_embed}, and the color difference $\Delta \mathbf{c}_{j}$ follows the same procedure. We exploit an additive and multiplicative trigonometric PE to weigh $\hat{\mathbf{f}}_{j}^{l}$,
\begin{equation}
\setlength{\abovedisplayskip}{3pt}
\setlength{\belowdisplayskip}{4pt}
\begin{aligned}
    &\mathbf{f}^{l,p}_{j} = \big(\hat{\mathbf{f}}_{j}^{l} + \operatorname{Emb}(\Delta \mathbf{p}_{j}; \mathbf{v})\big) \odot \operatorname{Emb}(\Delta \mathbf{p}_{j}; \mathbf{v}), \\
    &\mathbf{f}^{l,c}_{j} = \big(\hat{\mathbf{f}}_{j}^{l} + \operatorname{Emb}(\Delta \mathbf{c}_{j}; \mathbf{v})\big) \odot \operatorname{Emb}(\Delta \mathbf{c}_{j}; \mathbf{v}),
\end{aligned}
\label{equ:add_multiply}
\end{equation}
where $\odot$ represents element-wise multiplication. $\mathbf{v}=[v_1, ..., v_{2^{l}d}]$ are frequencies sampled from a Gaussian random distribution with variance $\delta $.
Then, the neighbor point $j$'s embedding is calculated via $ \alpha$-weighted summation,
\begin{equation}
\setlength{\abovedisplayskip}{3pt}
\setlength{\belowdisplayskip}{3pt}
\begin{aligned}
\label{weigh}
    \mathbf{f}^{l}_{j} = \alpha \cdot \mathbf{f}^{l,p}_{j} + (1 - \alpha) \cdot \mathbf{f}^{l,c}_{j} \quad \in \mathbb{R}^{2^{l} \times 6d},
\end{aligned}
\end{equation}
which implicitly embeds the representation of local shape into $\mathbf{f}^{l}_{j}$.
Finally, we use symmetric functions, maximum and average poolings, to compress the neighborhood information into the central point,
\begin{equation}
\setlength{\abovedisplayskip}{3pt}
\setlength{\belowdisplayskip}{1pt}
\begin{aligned}
    \mathbf{f}^{l} = \operatorname{MaxPool} ( \mathbf{f}^{l}_{j}) + \operatorname{AvgPool} ( \mathbf{f}^{l}_{j}),
\end{aligned}
\label{equ:local_pooling}
\end{equation}
where $\mathbf{f}^{l}$ is the representation of center point $\mathbf{p}$ in the $l$-th layer, which is an integrated representation of the local neighborhood.
We acquire hierarchical features $\{\mathbf{f}^{1}, \mathbf{f}^{2}, \mathbf{f}^{3} \}$ from the three local embedding layers and subsequently feed them into the upsampling layers.

\paragraph{\textit{Upsampling.}}
After all three local embedding layers, the encoder progressively upsamples the hierarchical features to the input point number for obtaining the final point-level features. As shown in Fig. \ref{fig:framework}, total three upsampling layers are involved. For the layer $l$, $l \in \left \{4, 5, 6 \right \}$, we first interpolate the central point embedding via the weighted sum of neighbor point embeddings,
\begin{equation}
\setlength{\abovedisplayskip}{3pt}
\setlength{\belowdisplayskip}{2pt}
\begin{aligned}
    \hat{\mathbf{f}}^{l} = \omega_{j} \sum_{j \in \mathcal{N}_{p}} \mathbf{f}^{l-1}_{j}, 
\end{aligned}
\end{equation}
where the weight $\omega_j$ is the inverse of the distance between the central point and the neighbor point $j$. Then, we concatenate the corresponding local embedding $\mathbf{f}^{6-l}$ and $\hat{\mathbf{f}}^{l}$ to obtain the output of upsampling layer $l$ as
\begin{equation}
\begin{aligned}
\mathbf{f}^{l} = \operatorname{Concat}(\mathbf{f}^{6-l}, \hat{\mathbf{f}}^{l}), l \in \{ 4, 5, 6\}.
\end{aligned}
\end{equation}
After the three upsampling layers, we obtain the final representation for each point, denoted as $\mathbf{f} \in \mathbb{R}^{D}$, where $D =(2^{0}+...+2^{3})\times 6d$. In this way, the training-free encoder produces the point-level embeddings of the input point cloud, denoted as $\mathbf{F}=\{ \mathbf{f}_{m} \} ^{M}_{m=1}$, $\mathbf{F} \in \mathbb{R}^{M \times D}$.

\paragraph{} By stacking the three types of layers, the training-free encoder can extract discriminative shape embeddings without introducing learnable parameters, thus it eliminates tedious pre-training and episodic training steps and substantially simplifies the few-shot pipeline. This advantage allows TFS3D to save significant time and resources and alleviate domain gaps caused by disparate training and testing classes.

\subsection{Similarity-based Segmentation}

For $N$-way-$K$-shot tasks, we utilize such a training-free encoder to respectively extract support-set features, denoted as $\mathbf{F}^{S} \in \mathbb{R}^{N \times K \times M \times D}$, and query-set features, $\mathbf{F}^{Q}\in \mathbb{R}^{Q \times M \times D}$, where $Q$ is the number of query samples. We also denote the support-set labels as $\mathbf{L}^{S} \in \mathbb{R}^{N \times K \times M}$. Then, we conduct a simple similarity-based segmentation.
We employ masked average pooling on support-set features to produce the prototypes of $N+1$ classes, denoted as $\mathbf{F}^{P} \in \mathbb{R}^{ (N+1)\times D}$. Mathematically, the prototype of class $n$, $\mathbf{F}^{P}_{n}$, is calculated via
\begin{equation}
\setlength{\abovedisplayskip}{5pt}
\setlength{\belowdisplayskip}{5pt}
\begin{aligned}
    \mathbf{F}^{P}_{n} = \frac{\sum_{k,m}\mathbf{F}_{n,k,m}^{S} \mathbf{1}(\mathbf{L}_{n,k,m}^{S}=n)}{\sum_{k,m} \mathbf{1}(\mathbf{L}_{n,k,m}^{S}=n)},
\end{aligned}
\end{equation}
where $n\in \{0,..., N\}, k\in \{1,..., K\}, m\in \{1,..., M\}$ and $\mathbf{F}_{n, k, m}^{S}\in \mathbb{R}^{D}$ represents the $m$-th point's embedding in the $k$-th support sample of the $n$-th category. $\mathbf{1}(\cdot)$ is a binary label indicator that outputs 1 when the input variable is true. 
After that, the cosine similarity is calculated between normalized query-set features and prototypes, 
\begin{equation}
\setlength{\abovedisplayskip}{3pt}
\setlength{\belowdisplayskip}{3pt}
\begin{aligned}
    \mathbf{S}_{cos} = \mathbf{F}^{Q}\mathbf{F}^{P}{}^{\top} \in \mathbb{R}^{Q\times M \times (N+1)},
\end{aligned}
\end{equation}
which represents the similarity between each point in the query set and $N+1$ prototypes.
Finally, we weight and integrate the one-hot labels of the prototypes, $\mathbf{L}^{P} \in \mathbb{R}^{(N+1) \times (N+1)}$, to achieve the final prediction, 
\begin{equation}
\setlength{\abovedisplayskip}{3pt}
\setlength{\belowdisplayskip}{3pt}
\begin{aligned}
    \mathrm{logits} = \varphi(\mathbf{S}_{cos} \mathbf{L}^{P})\ \in \mathbb{R}^{Q\times M \times (N+1)},
\end{aligned}
\end{equation}
where $\varphi(x)=\exp(-\gamma(1-x))$ acts as an activation function and $\gamma$ is a scaling factor~\cite{zhang2021tip,zhu2023not}.
In this process, the more similar prototypes contribute more to the final logits. 
By this segmentation head, the entire TFS3D framework can be purely non-parametric and training-free, thus achieving superior efficiency.

\begin{figure}[t]
\centering
\includegraphics[width=0.48\textwidth]{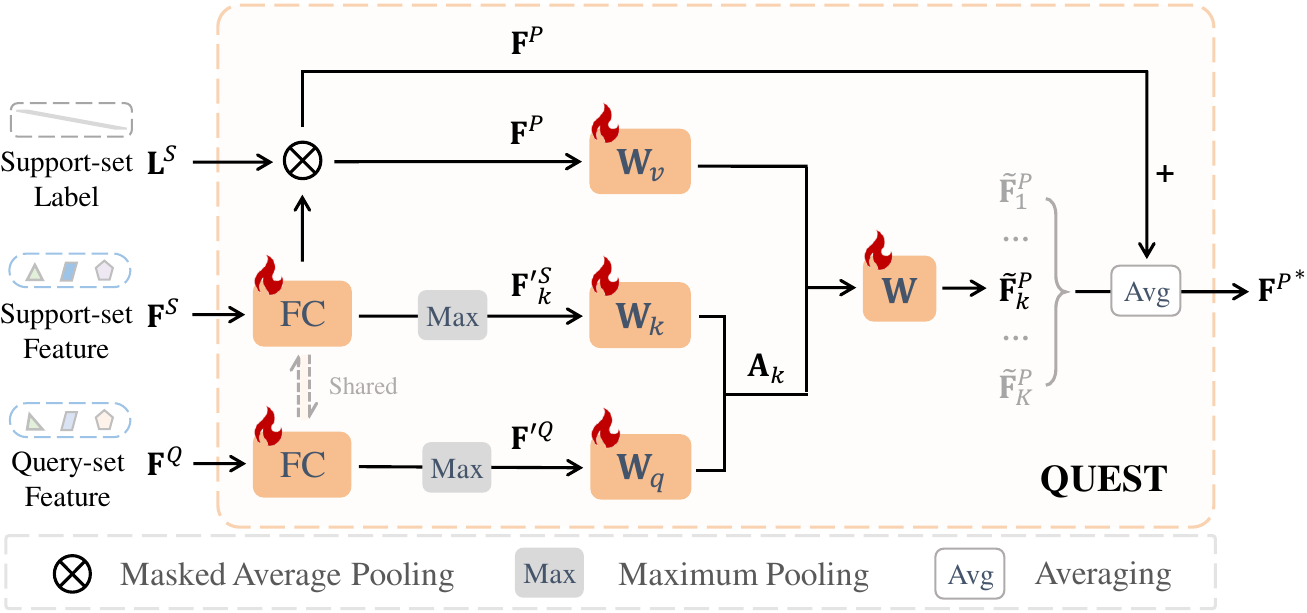}
\vspace{-0.4cm}
\caption{\textbf{Details of QUEST in TFS3D-T.} QUEST finally outputs adjusted prototypes $\mathbf{F}^{P}{}^{*}$.}
\label{fig:TFS3D-T}
\vspace{-0.2cm}
\end{figure}

\section{Training-based TFS3D-T}

To further achieve better segmentation performance, we propose a training-based variant, TFS3D-T. 
For parameter efficiency, TFS3D-T inherits the training-free 3D encoder of TFS3D to extract point cloud features. 
An obvious problem in few-shot learning is that the small-scale support set might fail to represent the true distribution of each category, which causes domain shift between support and query sets. This problem exists in kinds of 2D/3D tasks~\cite{yu2022mttrans,jaritz2020xmuda}. In our cases, the domain gap inevitably leads to biased prototypical learning. Inspired by \cite{he2023prototype}, we propose a QUEry-Support Transferring attention (QUEST) module to mitigate such bias problem, which transfers the prototypes from the support-set to the query-set domain. QUEST follows the training-free encoder and adjusts the prototypes based on the query-support interaction. After QUEST, we still adopt the similarity-based segmentation head of TFS3D to achieve the final prediction.

The detailed structure of QUEST is shown in Fig. \ref{fig:TFS3D-T}. We first leverage a shared fully connected (FC) layer to extract discriminative representations from support-set and query-set features. 
Then, We conduct local maximum pooling to obtain the statistics of each feature channel among the $M$ points in the support and query point clouds,
\begin{equation}
\setlength{\abovedisplayskip}{3pt}
\setlength{\belowdisplayskip}{3pt}
\begin{aligned}
    &\mathbf{F}'^{S} = \operatorname{MaxPool} (\mathbf{F}^{S}) \quad \in \mathbb{R}^{(N+1) \times K \times M' \times D}, \\
    &\mathbf{F}'^{Q} = \operatorname{MaxPool} (\mathbf{F}^{Q}) \quad \in \mathbb{R}^{Q \times M' \times D},
\end{aligned}
\label{equ:QUEST_maxpool}
\end{equation}
where the kernel size and stride of the pooling operation are hyperparameters. Eq. \ref{equ:QUEST_maxpool} produces $M'$ statistics from the $M$ points to manifest the distribution of each feature channel.
On top of this, QUEST bridges the support and query data via cross-attention. The $\mathbf{Q}$, $\mathbf{K}$, and $\mathbf{V}$ of the attention are,
\begin{equation}
\setlength{\abovedisplayskip}{4pt}
\setlength{\belowdisplayskip}{4pt}
\begin{aligned}
    \mathbf{Q} = {\mathbf{F}'^{Q}}{}^\top \mathbf{W}_{q}, 
    \mathbf{K}_{k} = {\mathbf{F}'^{S}_{k}}{}^\top \mathbf{W}_{k},
    \mathbf{V} = {\mathbf{F}^{P}}{}^{\top} \mathbf{W}_{v},
    \label{eq:qkv}
\end{aligned}
\end{equation}
where $\mathbf{W}_{q}$, $\mathbf{W}_{k}$ $\in \mathbb{R}^{M'\times M'}$ and $\mathbf{W}_{v} \in \mathbb{R}^{D\times D}$ are learnable projection matrices. $\mathbf{F}'^{S}_k, k \in \left \{ 1, ..., K\right \}$ indicates the contribution of the $k$-th shot from the support set. Next, we adjust each prototype channel to the query-set feature channel distribution. The attention matrix is
\begin{equation}
\setlength{\abovedisplayskip}{3pt}
\setlength{\belowdisplayskip}{3pt}
\begin{aligned}
    \mathbf{A}_{k} = \operatorname{softmax}(\frac{\mathbf{Q}^\top \cdot \mathbf{K}_{k}}{\sqrt{D}}) \quad \in \mathbb{R}^{Q \times (N+1) \times D \times D}.
    \label{eq:attn}
\end{aligned}
\end{equation}
The prototype channels are adjusted by matrix multiplication between $\mathbf{A}_{k}$ and the transpose of $\mathbf{V}$. We denote the adjusted prototype as $\tilde{\mathbf{F}}^{P}_{k}$, then it is calculated via
\begin{equation}
\setlength{\abovedisplayskip}{3pt}
\setlength{\belowdisplayskip}{3pt}
\begin{aligned}
\tilde{\mathbf{F}}^{P}_{k} = \mathbf{W} (\mathbf{A}_{k} \cdot \mathbf{V}^\top)^\top, 
\label{eq:get_FPk}
\end{aligned}
\end{equation}
where $\mathbf{W}$ is a learnable matrix. We integrate the adjusted and original prototypes to obtain final prototypes $\mathbf{F}^{P}{}^{*}= \mathbf{F}^{P} + \sum_{k=1}^{K}\tilde{\mathbf{F}}_{k}^{P}$, 
where the original prototypes $\mathbf{F}^{P}$ is obtained via masked average pooling and we average all $K$-shot adjusted prototypes. 

After QUEST, we utilize the same similarity matching scheme as TFS3D to segment the query set. As the 3D encoder is training-free, we do not require pre-training and only require episodic training to learn the QUEST module. During training, we adopt cross-entropy loss to supervise the optimization of QUEST. 

\begin{table*}[t!]
\centering
\begin{adjustbox}{width=0.97\linewidth}
\begin{tabular}{l|c||ccc|ccc||ccc|ccc}
\toprule
\multirow{3}{*}{\textbf{Method}} & \multirow{3}{*}{\textbf{Parameters}}
	& \multicolumn{6}{c||}{\textbf{Two-way}} & \multicolumn{6}{c}{\textbf{Three-way}}\\
	\cline{3-14}
	& & \multicolumn{3}{c|}{\textbf{One-shot}} & \multicolumn{3}{c||}{\textbf{Five-shot}} & 
	\multicolumn{3}{c|}{\textbf{One-shot}} & \multicolumn{3}{c}{\textbf{Five-shot}} \\
	\cline{3-14}
	& & $S_0$ & $S_1$ & $Avg$ & $S_0$ & $S_1$ & $Avg$ & $S_0$ & $S_1$ & $Avg$ & $S_0$ & $S_1$ & $Avg$ \\
	\midrule

    \rowcolor{purple!5} Point-NN & 0.00~M & 42.12 & 42.62 & 42.37 & 51.91 & 49.35 & 50.63 & 38.00 & 36.21 & 37.10 & 45.91 & 43.44 & 44.67\\

    \rowcolor{purple!5} TFS3D & 0.00~M & {51.42} & {54.94} & 53.18 & {56.09} & {58.49} & 57.29 & 42.79 & 45.99 & 44.39 & 48.28 & 50.77 & 49.52 \\

    \rowcolor{purple!5} \textit{\small{Improvement}} & - & \textcolor{blue}{\small{+9.3}} & \textcolor{blue}{\small{+12.32}} & \textcolor{blue}{\small{+10.81}} & \textcolor{blue}{\small{+4.18}} & \textcolor{blue}{\small{+9.14}} & \textcolor{blue}{\small{+6.66}} & \textcolor{blue}{\small{+4.79}} & \textcolor{blue}{\small{+9.78}} & \textcolor{blue}{\small{+7.29}} & \textcolor{blue}{\small{+2.37}} & \textcolor{blue}{\small{+7.33}} & \textcolor{blue}{\small{+4.85}} \\
	\midrule
    
    DGCNN & {0.62~M} & 36.34 & 38.79 & 37.57 & 56.49 & 56.99 & 56.74 & 30.05 & 32.19 & 31.12 & 46.88 & 47.57 & 47.23 \\
 
	ProtoNet & 0.27~M & 48.39 & 49.98 & 49.19 & 57.34 & 63.22 & 60.28 & 40.81 & 45.07 & 42.94 & 49.05 & 53.42 & 51.24 \\ 
	
	MPTI & 0.29~M & 52.27 & 51.48 & 51.88 & 58.93 & 60.56 & 59.75 & 44.27 & 46.92 & 45.60 & 51.74 & 48.57 & 50.16 \\ 
	
	AttMPTI & 0.37~M & 53.77 & 55.94 & 54.86 & 61.67 & 67.02 & 64.35 & 45.18 & 49.27 & 47.23 & 54.92 & 56.79 & 55.86 \\ 

    BFG & - & 55.60 & 55.98 & 55.79 & 63.71 & 66.62 & 65.17 & 46.18 & 48.36 & 47.27 & 55.05 & 57.80 & 56.43 \\ 
 
    2CBR & 0.35~M & 55.89 & 61.99 & 58.94 & 63.55 & 67.51 & 65.53 & 46.51 & 53.91 & 50.21 & 55.51 & 58.07 & 56.79 \\ 
    
	PAP3D & 2.45~M & 59.45 & 66.08 & 62.76 & 65.40 & 70.30 & 67.85 & 48.99 & 56.57 & 52.78 & 61.27 & 60.81 & 61.04 \\
	\midrule

    TFS3D-T & \textbf{0.25~M}& \textbf{70.34} & \textbf{74.09} & \textbf{72.21} & \textbf{70.62} & \textbf{74.50} & \textbf{72.56} & \textbf{61.54} & \textbf{63.39} & \textbf{62.46} & \textbf{62.05} & \textbf{67.82} & \textbf{64.93}  \\

    \textit{\small{Improvement}} & - & \textcolor{blue}{\small{+10.89}} & \textcolor{blue}{\small{+8.01}} & \textcolor{blue}{\small{+9.45}} & \textcolor{blue}{\small{+5.22}} & \textcolor{blue}{\small{+4.20}} & \textcolor{blue}{\small{+4.71}} & \textcolor{blue}{\small{+12.55}} & \textcolor{blue}{\small{+6.82}} & \textcolor{blue}{\small{+9.68}} & \textcolor{blue}{\small{+0.78}} & \textcolor{blue}{\small{+7.01}} & \textcolor{blue}{\small{+3.89}} \\
    
    \bottomrule
\end{tabular}
\end{adjustbox}
\caption{\textbf{Few-shot Results (\%) on S3DIS.} $S_i$ denotes the split $i$ is used for testing, and $Avg$ is their average mIoU. The shaded rows represent training-free methods. `Parameters' represents the total number of learnable parameters of each method.}
\label{table:s3dis_iou}
\end{table*}

\begin{table*}[t]
\centering
\begin{adjustbox}{width=1.0\linewidth}
\begin{tabular}{l|c||ccc|ccc||ccc|ccc}
\toprule
\multirow{3}{*}{\textbf{Method}}
	& \multirow{3}{*}{\textbf{Parameters}} & \multicolumn{6}{c||}{\textbf{Two-way}} & \multicolumn{6}{c}{\textbf{Three-way}}  \\
	\cline{3-14}
	& & \multicolumn{3}{c|}{\textbf{One-shot}} & \multicolumn{3}{c||}{\textbf{Five-shot}} &
	\multicolumn{3}{c|}{\textbf{One-shot}} & \multicolumn{3}{c}{\textbf{Five-shot}} \\
	\cline{3-14}
	& & $S_0$ & $S_1$ & $Avg$ & $S_0$ & $S_1$ & $Avg$ & $S_0$ & $S_1$ & $Avg$ & $S_0$ & $S_1$ & $Avg$ \\
	\midrule

    \rowcolor{purple!5} Point-NN & 0.00~M & 35.32 & 34.13 & 34.72 & 44.12 & 43.40 & 43.76 & 26.81 & 25.20 & 26.00 & 34.07 & 32.16 & 33.11 \\
    \rowcolor{purple!5} TFS3D & 0.00~M & {40.40} & {37.53} & 38.96 & {46.55} & {43.04} & 44.79 & 30.26 & 26.33 & 28.29 & 36.07 & 31.47 & 33.77 \\

    \rowcolor{purple!5} \textit{\small{Improvement}} & 
    - & \textcolor{blue}{\small{+5.08}} & \textcolor{blue}{\small{+3.40}} & \textcolor{blue}{\small{+4.24}} & \textcolor{blue}{\small{+2.43}} & \textcolor{blue}{\small{-0.36}} &\textcolor{blue}{\small{+1.03}} & \textcolor{blue}{\small{+3.45}} & \textcolor{blue}{\small{+1.13}} & \textcolor{blue}{\small{+2.29}} & \textcolor{blue}{\small{+2.00}} & \textcolor{blue}{\small{-0.69}} & \textcolor{blue}{\small{+0.66}}  \\
	\midrule
    
    DGCNN & {1.43~M} & 31.55 & 28.94 & 30.25 & 42.71 & 37.24 & 39.98 & 23.99 & 19.10 & 21.55 & 34.93 & 28.10 & 31.52 \\
 
	ProtoNet & 0.27~M & 33.92 & 30.95 & 32.44 & 45.34 & 42.01 & 43.68 & 28.47 & 26.13 & 27.30 & 37.36 & 34.98 & 36.17 \\ 
    
    MPTI & 0.29~M & 39.27 & 36.14 & 37.71 & 46.90 & 43.59 & 45.25 & 29.96 & 27.26 & 28.61 & 38.14 & 34.36 & 36.25 \\ 
    
    AttMPTI & 0.37~M & 42.55 & 40.83 & 41.69 & 54.00 & 50.32 & 52.16 & 35.23 & 30.72 & 32.98 & 46.74 & 40.80 & 43.77 \\ 
    
    BFG & - & 42.15 & 40.52 & 41.34 & 51.23 & 49.39 & 50.31 & 34.12 & 31.98 & 33.05 & 46.25 & 41.38 & 43.82 \\
    
    2CBR & 0.35~M & 50.73 & 47.66 & 49.20 & 52.35 & 47.14 & 49.75 & 47.00 & 46.36 & 46.68 & 45.06 & 39.47 & 42.27 \\ 
    
    PAP3D & 2.45~M & 57.08 & 55.94 & 56.51 & 64.55 & 59.64 & 62.10 & 55.27 & 55.60 & 55.44 & 59.02 & 53.16 & 56.09 \\
	\midrule
    
    TFS3D-T & \textbf{0.25~M} & \textbf{78.91} & \textbf{78.73} & \textbf{78.82} & \textbf{79.39} & \textbf{80.23} & \textbf{79.81} & \textbf{72.09} & \textbf{69.41} & \textbf{70.75} & \textbf{72.73} & \textbf{72.38} & \textbf{72.55} \\

    \textit{\small{Improvement}} & 
    - & \textcolor{blue}{\small{+21.83}} & \textcolor{blue}{\small{+22.79}} & \textcolor{blue}{\small{+22.31}} & \textcolor{blue}{\small{+14.84}} & \textcolor{blue}{\small{+20.59}} &\textcolor{blue}{\small{+17.71}} & \textcolor{blue}{\small{+16.82}} & \textcolor{blue}{\small{+13.81}} & \textcolor{blue}{\small{+15.35}} & \textcolor{blue}{\small{+13.71}} & \textcolor{blue}{\small{+19.22}} & \textcolor{blue}{\small{+16.46}}  \\
    
\bottomrule
\end{tabular}
\end{adjustbox}
\caption{\textbf{Few-shot Results (\%) on ScanNet.} $S_i$ denotes the split $i$ is used for testing, and $Avg$ is their average mIoU. The shaded rows represent training-free methods. `Parameters' represents the total number of learnable parameters of each method.}
\label{table:scannet_iou}
\vspace{-0.3cm}
\end{table*}

\begin{table}[t!]
\vspace{0.2cm}
\centering
\begin{adjustbox}{width=0.997\linewidth}
	\begin{tabular}{lccccc}
	\toprule
		\makecell*[c]{\textbf{Method}} & \textbf{mIoU} & \makecell*[c]{\textbf{Pre-train}} & \textbf{Parameters} &\makecell*[c]{\textbf{Train}\\\textbf{Time}} &\makecell*[c]{\textbf{Test}\\\textbf{Speed}}\\
		\cmidrule(lr){1-1} \cmidrule(lr){2-2} \cmidrule(lr){3-3} \cmidrule(lr){4-4} \cmidrule(lr){5-5} \cmidrule(lr){6-6}
	    attMPTI & 53.77 & $\checkmark$ & 0.37~M & 9.5~h & 5 \\
        2CBR & 55.89 & $\checkmark$ & 0.35~M & 6.2~h & \textbf{66} \\
        PAP3D & 59.45 & $\checkmark$ & 2.45~M & 4.7~h & 60 \\

        \cmidrule(lr){1-6}
        \rowcolor{purple!5} TFS3D & 51.42 & \ding{55} & 0~M & 0~h & 22 \\

        TFS3D-T* & \textbf{62.51} & \ding{55} & \textbf{0.19~M} & \textbf{0.4~h} & 64 \\
        TFS3D-T & \textbf{70.34} & \ding{55} & \textbf{0.25~M} & \textbf{0.5~h} & 48 \\
    
	\bottomrule
	\end{tabular}
\end{adjustbox}
\caption{\textbf{Performance (\%) and Efficiency Comparison on S3DIS}. Train Time and Test Speed (episodes/second) are tested on one NVIDIA A6000 GPU. We calculate the total time of pre-training and episodic training. We report the accuracy under 2-way-1-shot settings on $S_0$ split. We simplify TFS3D-T by setting the initial frequency number to $d=5$, denoted as `TFS3D-T*', which achieves higher test speed.}
\label{table:efficiency_comparison}
\vspace{-0.5cm}
\end{table}

\section{Experiments}

In this section, we first introduce the datasets and implementation details. Then we report the experimental results in comparison with existing approaches. At last, we present ablation studies to verify our effectiveness.

\subsection{Experimental Details}

\paragraph{Datasets.} 
We validate our method using two public 3D datasets, S3DIS~\cite{armeni20163d} and ScanNet~\cite{dai2017scannet}. 
Due to the large scale of original scenes, we adopt the data pre-processing in \cite{zhao2021few, he2023prototype} and partition them into small blocks. Then S3DIS and ScanNet contain 7,547 and 36,350 blocks, respectively. $M=2048$ points are randomly sampled from each block. For each dataset, we generate a training class set $C_{train}$ and a testing class set $C_{test}$ that have no overlap. We use $C_{train}$ for episodic training and $C_{test}$ for testing, performing cross-validation for each dataset. For $N$-way-$K$-shot test episodes, we iterate over all combinations of $N$ classes from $C_{test}$, sampling 100 episodes for each combination.

\paragraph{Basic Settings.} The TFS3D encoder is frozen in all experiments. For few-shot settings, we experiment under 2/3-way-1/5-shot settings respectively, following \cite{zhao2021few, mao2022bidirectional, he2023prototype}. For performance, we adopt the mean Intersection over Union (mIoU) as evaluation criteria. mIoU is computed by averaging the IoU scores across all unseen classes $C_{test}$. We provide more detailed settings in supplementary materials.


\subsection{Analysis}

\paragraph{Baselines} To validate the performance of our method, we compare it with two types of methods. First, we consider training-based 2D/3D few-shot segmentation methods, including DGCNN~\cite{shaban2017one}, ProtoNet~\cite{garcia2017few}, MTPI~\cite{zhao2021few}, AttMPTI~\cite{zhao2021few}, BFG~\cite{mao2022bidirectional}, 2CBR~\cite{zhu2023cross}, and PAP3D~\cite{he2023prototype}. Second, we re-implement the only training-free and non-parametric model, Point-NN~\cite{zhang2023parameter}, under our settings. In Tab. \ref{table:s3dis_iou} and \ref{table:scannet_iou}, we report the results of TFS3D and TFS3D-T on S3DIS and ScanNet datasets.

\paragraph{Performance.} For the training-free TFS3D, we compare it to Point-NN and observe significantly improved performance on the S3DIS dataset. In addition, TFS3D even outperforms some training-based methods, such as DGCNN and ProtoNet without any training. For training-based TFS3D-T, its results significantly outperform previous SOTA mIoU across all four few-shot tasks on 2 datasets. On S3DIS, it achieves an average improvement of \textbf{+6.93\%} across four tasks. On the more challenging ScanNet dataset, we observe significant improvements, \textbf{+17.96\%} average mIoU across four tasks compared to the currently best-performing PAP3D. These enhancements demonstrate that our method can better alleviate domain gaps between seen and unseen classes and address the prototype bias problem. This also indicates our training-free encoder can extract discriminative and general knowledge for 3D shapes. For parameter number, we only use 0.25M parameters, the least among existing methods and \textbf{-90\%} less than PAP3D. 

\paragraph{Efficiency.} In Tab. \ref{table:efficiency_comparison}, we compare the time efficiency with existing works.
For TFS3D, no training is required, thus it achieves few-shot segmentation with minimal resource and time consumption. For training-based TFS3D-T, we require only episodic training without pre-training, thus our model greatly reduces training time by more than \textbf{-90\%} compared to existing methods. Both TFS3D and TFS3D-T efficiently simplify the few-shot pipeline by discarding the pre-training step. However, the proposed multi-layer 3D encoder of TFS3D may lead to increased inference time. We can mitigate this issue by reducing the feature dimension. By setting the initial frequency number to $d=5$, denoted as `TFS3D-T*' in Tab. \ref{table:efficiency_comparison}, we achieve high speed without significantly sacrificing performance. Overall, our model demonstrates advantages in both training time and inference speed.

\begin{figure}[t!]
\centering
\includegraphics[width=0.457\textwidth]{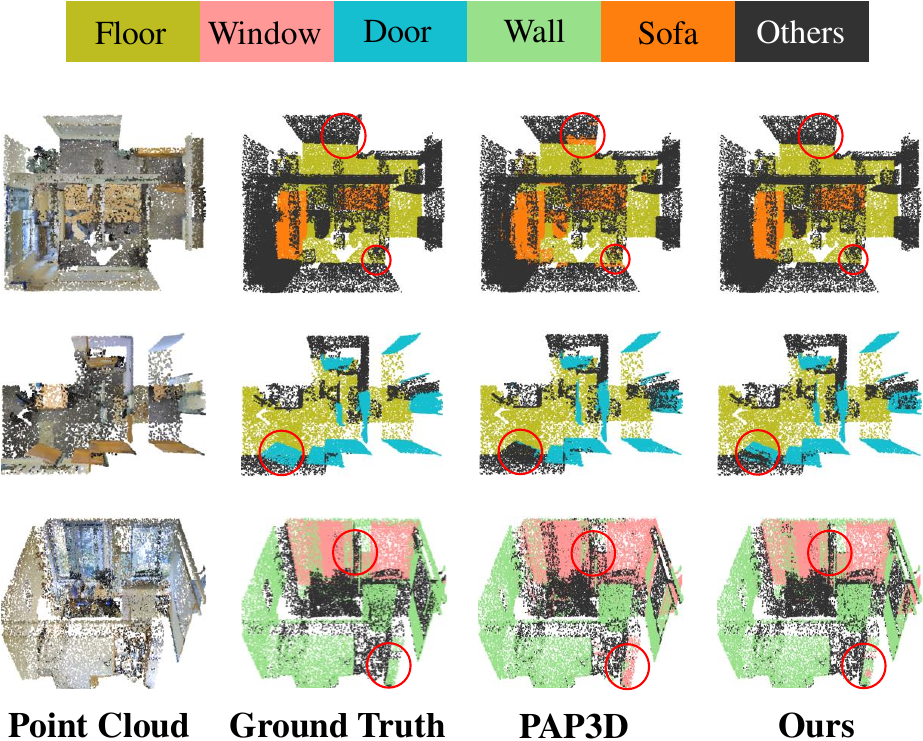}
\vspace{-0.1cm}
\caption{\textbf{Visualization of Results} on S3DIS dataset. We compare TFS3D-T's prediction with the SOTA PAP3D.}
\label{fig:eg_s3dis}
\vspace{-0.07cm}
\end{figure}

\begin{table}[t]
\centering
\begin{adjustbox}{width=0.99\linewidth}
\begin{tabular}{l|ccc|ccc}
\toprule
    \multirow{2}{*}{\textbf{Local PE}} & \multicolumn{3}{c|}{\textbf{TFS3D}} & \multicolumn{3}{c}{\textbf{TFS3D-T}} \\ \cline{2-7}

    & $S_0$ & $S_1$ & $Avg$ & $S_0$ & $S_1$ & $Avg$ \\  \midrule
    
    w/o & 45.71 & 47.85 & 46.78 & 58.34 & 61.33 & 59.83 \\
    
    add & 47.34 & 51.26 & 48.30 & 66.73 & 71.79 & 69.26 \\

    multiply & 49.84 & 53.12 & 51.48 & \textbf{71.65} & 70.60 & 71.12 \\

    add+multiply & \textbf{51.42} & \textbf{54.94} & \textbf{53.18} & 70.34 & \textbf{74.09} & \textbf{72.21} \\
\bottomrule
\end{tabular}
\end{adjustbox}
\vspace{-0.1cm}
\caption{\textbf{Ablation for PE Weighing Design} of Eq. \ref{equ:add_multiply} in local embedding layers. We compare the additive and multiplicative PEs and their combinations.}
\label{table:ablation_PE_weigh}
\vspace{-0.1cm}
\end{table}

\begin{table}[t!]
\centering
\begin{adjustbox}{width=0.88\linewidth}
\begin{tabular}{l|ccc|ccc}
\toprule
    \multirow{2}{*}{\textbf{}} & \multicolumn{3}{c|}{\textbf{Coordinate}} & \multicolumn{3}{c}{\textbf{+ Color}} \\ \cline{2-7}

    & $S_0$ & $S_1$ & $Avg$ & $S_0$ & $S_1$ & $Avg$ \\  \midrule

    TFS3D & 48.81 & 52.31 & 50.57 & \textbf{51.42} & \textbf{54.94} & \textbf{53.18} \\
\bottomrule
\end{tabular}
\end{adjustbox}
\vspace{-0.1cm}
\caption{\textbf{Ablation for Coordinate and Color Information}. }
\label{table:ablation_position_color}
\vspace{-0.3cm}
\end{table}

\paragraph{Visualization.} We present several qualitative results of 2-way-1-shot tasks in Fig. \ref{fig:eg_s3dis} and Fig. \ref{fig:eg_scannet}. TFS3D-T achieves better segmentation than the existing SOTA, PAP3D~\cite{he2023prototype}. TFS3D-T demonstrates effectiveness not only for common furnitures, such as sofas and chairs, but also for less frequent categories like bathtubs (Fig. \ref{fig:eg_scannet}, row 2). These results demonstrate the effectiveness of TFS3D-T.

\subsection{Ablation Study}

We conduct a series of ablation experiments to reveal the roles of different designs or modules. By default, we perform 2-way-1-shot experiments on the S3DIS dataset, including both $S_0$ and $S_1$ splits.

\paragraph{Ablation for TFS3D.} We first investigate the different designs of PEs in local embedding layers to weigh neighbor points in Tab. \ref{table:ablation_PE_weigh}. We compare additive and multiplicative PEs used in Eq. \ref{equ:add_multiply} and conclude that combining additive and multiplicative PEs yields the best performance. In Tab. \ref{table:ablation_position_color}, we investigate the role of the coordinate and color information and find combining both results in a +2.61\% improvement compared to using only coordinate information. 

\begin{figure}[t]
\centering

\includegraphics[width=0.466\textwidth]{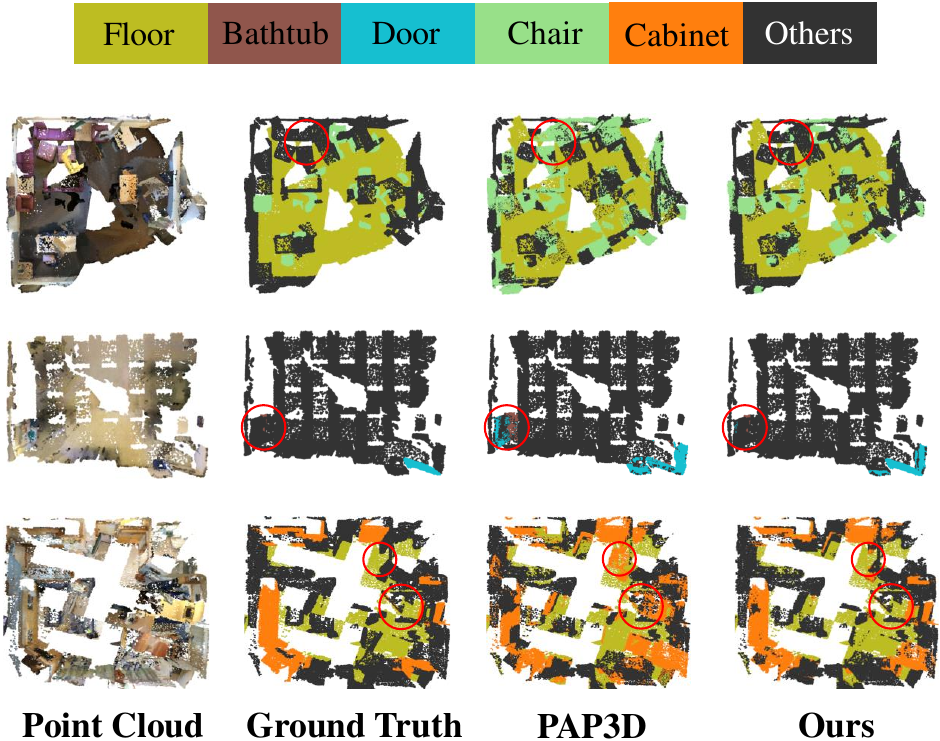}
\vspace{-0.1cm}
\caption{\textbf{Visualization of Results} on ScanNet dataset. We compare TFS3D-T's results with the SOTA PAP3D model.}
\label{fig:eg_scannet}
\vspace{-0.08cm}
\end{figure}

\begin{table}[t!]
\centering
\begin{adjustbox}{width=0.93\linewidth}
\begin{tabular}{l|ccc|ccc}
\toprule

    \multirow{2}{*}{\textbf{QUEST}} & \multicolumn{3}{c|}{\textbf{Two Way}} & \multicolumn{3}{c}{\textbf{Three Way}} \\ \cline{2-7}

    & $S_0$ & $S_1$ & $Avg$ & $S_0$ & $S_1$ & $Avg$ \\  \midrule
    
    w/o & {51.42} & {54.94} & {53.18} & 42.79 & 45.99 & 44.39 \\
    
    FC & 60.74 & 65.75 & 63.25 & 57.93 & 58.81 & 58.37 \\

    Attention & 59.47 & 72.21 & 65.84 & {59.46} & 60.49 & 60.12 \\

    QUEST & \textbf{70.34} & \textbf{74.09} & \textbf{72.21} & \textbf{61.54} & \textbf{63.39} & \textbf{62.46} \\
\bottomrule
\end{tabular}
\end{adjustbox}
\vspace{-0.1cm}
\caption{\textbf{Ablation for QUEST} in TFS3D-T. We report the results (\%) under 2/3-way-1-shot settings. `FC' represents fully connected layers and 'Attention' for cross-attention.}
\label{table:ablation_QUEST}
\vspace{-0.1cm}
\end{table}

\begin{table}[t!]
\centering
\begin{adjustbox}{width=0.9\linewidth}
\begin{tabular}{l|ccc|ccc}
\toprule

    \multirow{2}{*}{\textbf{Encoder}} & \multicolumn{3}{c|}{\textbf{Two Way}} & \multicolumn{3}{c}{\textbf{Three Way}} \\ \cline{2-7}

    & $S_0$ & $S_1$ & $Avg$ & $S_0$ & $S_1$ & $Avg$ \\  \midrule
    
    DGCNN & 62.48 & 69.98 & 66.23 & {55.33} & 56.87 & 56.06 \\

    TFS3D & \textbf{70.34} & \textbf{74.09} & \textbf{72.21} & \textbf{61.54} & \textbf{63.39} & \textbf{62.46} \\
\bottomrule
\end{tabular}
\end{adjustbox}
\vspace{-0.1cm}
\caption{\textbf{Ablation for the 3D Encoder} on S3DIS. We report the results (\%) under 2/3-way-1-shot settings.}
\label{table:TFS3D-T_DGCNN}
\vspace{-0.3cm}
\end{table}

\paragraph{Ablation on TFS3D-T.} First, we investigate the effect of different 3D encoders on TFS3D-T. We substitute our training-free encoder with a pre-trained DGCNN (following \cite{he2023prototype}). From Tab. \ref{table:TFS3D-T_DGCNN}, we observe that using DGCNN results in a performance drop. This highlights the effectiveness of our encoder in capturing 3D geometrics. Then, we examine different designs of the QUEST module in Tab. \ref{table:ablation_QUEST}. We investigate the fully connected layer and the cross-attention operation. We observe that using only attention achieves lower performance than employing both. However, incorporating the fully connected layer with QUEST allows for better learning of the query-support interaction. 

\section{Conclusion}

We propose a training-free framework, TFS3D, and a training-based variant, TFS3D-T, for few-shot point cloud semantic segmentation. Based on trigonometric PEs, TFS3D introduces no learnable parameters and discards any training. TFS3D achieves comparable performance with training-based methods with minimal resource consumption. Furthermore, TFS3D-T introduces QUEST module to overcome prototype bias and achieves remarkable improvement compared to existing approaches. Importantly, TFS3D-T does not require pre-training, thus simplifying the pipeline of 3D few-shot segmentation and saving significant time. For future work, we will extend our training-free encoders and QUEST attention to other 3D applications, such as point cloud pre-training~\cite{zhang2022point,guo2023joint, yu2021pointbert, pang2022masked}, cross-modal learning~\cite{zhang2023learning, chen2023pimae, wang2022p2p, yan20222dpass} and open-world understanding~\cite{zhang2022pointclip, zhu2022pointclip, xue2022ulip, liu2023openshape}.

\bibliography{aaai24}

\newpage
\appendix
\twocolumn

The supplementary material of this work is provided below. We first present detailed descriptions of the experimental setup and hyperparameters, along with additional framework details. Then, we provide further experiment results and ablation studies.

\section{Experimental Setup}

\paragraph{Dataset Split} 
S3DIS consists of 272 room point clouds from three different buildings with distinct architectural styles and appearances. We exclude the background clutter class and focus on 12 explicit semantic classes. ScanNet comprises 1,513 point cloud scans from 707 indoor scenes, with 20 explicit semantic categories provided for segmentation. 
Tab. \ref{tab:data_split} lists the class names in the seen and unseen split of the S3DIS and ScanNet datasets.

\paragraph{Hyperparameters} In TFS3D, we set the frequency number in the initial embedding layer to $d=15$ and sample log-linear spaced frequencies $\mathbf{u}$ with $\theta=20$. The weight to integrate coordinate and color information in Equ. (2) and (5) of the main paper is set to $\alpha=0.8$. In local embedding layers, we sample frequencies $\mathbf{v}$ from a Gaussian random distribution, whose variance is set to $\delta=0.1$. We sample 8 neighbor points with $k$-NN to build the neighborhood of the center point for both local embedding and upsampling layers. In the similarity-based segmentation head, we set the scale factor $\gamma$ to 400. For TFS3D-T, we only train the fully connected layer and the QUEST module. We set the initial frequency number to $d=10$ and keep others the same as in TFS3D. In the QUEST module, we set the kernel size and stride of the local maximum pooling to 32. The fully connected layer in QUEST contains 2 linear projection operations, and each linear projection is followed by batch normalization and rectified linear activation functions. The detailed structure of the fully connected layer is: `(BN+ReLU) + (Linear+BN+ReLU) + (Linear+BN+ReLU)', where `BN', `ReLU', and `Linear' represent batch normalization, rectified linear activation, and linear projection, respectively.

\paragraph{Training Details} The proposed TFS3D and TFS3D-T are implemented using PyTorch. TFS3D-T is trained on a GForce A6000 GPU. The meta-training is performed directly on $C_{train}$ split, using AdamW optimizer ($\beta_{1} = 0.9, \beta_{2} = 0.999$) to update the QUEST module of TFS3D-T. The initial learning rate is set to 0.001, and halved every 7,000 iterations. In episodic training, each batch contains 1 episode, which includes a support set and a query set. The support set randomly selects $N$-way-$K$-shot point clouds and the query set randomly selects $Q=N$ samples.

\begin{table}[t]
\vspace{4pt}
\centering
\begin{adjustbox}{width=0.97\linewidth}
	\begin{tabular}{c|c|c}
	\toprule
	 & $S_0$ & $S_1$ \\
        \hline
		 \textbf{S3DIS} & \thead{beam, board, bookcase, \\ ceiling, chair, column} & \thead{door, floor, sofa, \\ table, wall, window} \\
		 \hline
         \textbf{ScanNet} & \thead{bathtub, bed, bookshelf, \\
        cabinet, chair, counter, \\ curtain, desk, door, \\floor} & \thead{other furniture, picture, \\refrigerator, show curtain,\\
        sink, sofa, table, \\toilet, wall, window} \\
		\specialrule{0em}{1pt}{1pt}
	\bottomrule
	\end{tabular}
\end{adjustbox}
\caption{\textbf{Seen and Unseen Classes Split} for S3DIS and ScanNet. We evenly distribute $S_0$ and $S_1$ splits.
}
\label{tab:data_split}
\vspace{-0.3cm}
\end{table}

\section{Additional Ablation Study}

We conduct additional ablation experiments to reveal the roles of different detailed designs. By default, we still perform 2-way-1-shot experiments on the S3DIS dataset.

\paragraph{Ablation on Coordinate and Color Information.} In Tab. \ref{table:ablation_position_color}, we exhibit more results of the investigation of the role that the positional and color information play. For TFS3D, both types of information are helpful for segmentation. However, in TFS3D-T, we observed that positional information alone can still achieve high performance.

\paragraph{Ablation on Frequency Distribution.} 

In Tab. \ref{table:ablation_distribution}, we compare the effects of different frequency distributions on local feature extraction. We adopt log-linear spaced frequencies in the initial trigonometric layer, and we observe variations in the ability of different frequency distributions to reveal local geometrics. We compare 3 additional distributions, Gaussian random, Laplacian random, and uniform distribution. By comparisons, we observe that the best performance is achieved when the initial embedding follows a log-linear distribution and the local embedding follows a Gaussian random distribution. This may be because the log-linear distribution covers a wider frequency range, allowing for more preservation of fine-grained initial information. 

\begin{table}[t!]
\centering
\begin{adjustbox}{width=0.99\linewidth}
\begin{tabular}{l|ccc|ccc}
\toprule
    \multirow{2}{*}{\textbf{mIoU}} & \multicolumn{3}{c|}{\textbf{Coordinate}} & \multicolumn{3}{c}{\textbf{+ Color}} \\ \cline{2-7}

    & $S_0$ & $S_1$ & $Avg$ & $S_0$ & $S_1$ & $Avg$ \\  \midrule

    TFS3D & 48.81 & 52.31 & 50.57 & \textbf{51.42} & \textbf{54.94} & \textbf{53.18} \\

    TFS3D-T & 65.68 & \textbf{74.98} & 70.33 & \textbf{70.34} & 74.09 & \textbf{72.21} \\
\bottomrule
\end{tabular}
\end{adjustbox}
\vspace{-0.1cm}
\caption{\textbf{Ablation for Coordinate and Color Information} on S3DIS. We report TFS3D and TES3D-T's results (\%).}
\label{table:ablation_position_color}
\end{table}

\begin{table}[t]
\centering
\begin{adjustbox}{width=0.98\linewidth}
\begin{tabular}{l|cccc|cc}
\toprule
    \multirow{2}{*}{\textbf{Initial}} & \multicolumn{4}{c|}{\textbf{Local Embedding}} & \multirow{2}{*}{\textbf{TFS3D}} & \multirow{2}{*}{\textbf{TFS3D-T}}\\ \cline{2-5}

    & Log & Uni & Gaus & Lap \\  \hline
    
    \multirow{4}{*}{{Log}} & $\checkmark$ & & & & 46.47 & 69.99 \\
    
    & & $\checkmark$ & & & 42.15 & 68.64 \\

    & & & $\checkmark$ & & \textbf{53.18} & \textbf{72.21} \\
    & & & & $\checkmark$ & 52.68 & 70.19 \\
\bottomrule
\end{tabular}
\end{adjustbox}
\caption{\textbf{Ablation Study for Frequency Distribution} on S3DIS. We report the average results (\%) of $S_0$ and $S_1$ split. The `Log', `Uni', `Gaus', and `Lap' represent the log-linear distribution, the uniform distribution, the Gaussian distribution, and the Laplacian distribution, respectively.}
\label{table:ablation_distribution}
\vspace{-0.3cm}
\end{table}

\paragraph{Ablation on Position Encoding Dimension.} In Tab. \ref{table:ablation_PEdim_TFS3D}, we examine the effect of the positional encoding dimension. We observe that 90 is the best choice of the dimension, and reducing the dimension has the potential to significantly impair performance.
In Tab. \ref{table:ablation_PEdim_TFS3D-T}, we investigate the impact of different dimensionality settings for the position encoding in TFS3D-T. We explore dimensions in $\{30, 36, 48, 60, 72, 90\}$. From the table, we observe that TFS3D-T achieves the best overall performance when the dimension is set to 60. In addition, reducing the dimension does not lead to significant performance degradation.

\paragraph{Hyperparameters in TFS3D.} \textbf{1) Initial Layer Parameter $\mathbf{\theta}$.} In the initial trigonometric layer, we utilized log-spaced frequencies $\mathbf{u}=[u_1, ..., u_d]$, where $u_{i}=\theta^{i/d}$ follows a log-linear distribution with the base number $\theta$. In Tab. \ref{table:ablation_theta}, we demonstrate the impact of different values of 
$\theta$ on training-free TFS3D and training-based TFS3D-T. We explore $\delta$'s value in $\{10, 20, 40, 60, 80, 100\}$. From the table, we observe that TFS3D is more sensitive to 
$\theta$, while TFS3D-T exhibits higher tolerance. 
\textbf{2) Local Layer Parameter $\delta$.} In local embedding layers, we utilize different frequencies $\mathbf{v}$ from the initial frequencies $\mathbf{u}$, where $\mathbf{v}=[v_1, ..., v_{2^{l}d}]$ are frequencies sampled from a Gaussian random distribution with variance $\delta$. Tab. \ref{table:ablation_delta} presents the influence of different $\delta$ values. We compare $\delta$'s value in $\{0.01, 0.05, 0.07, 0.1, 0.2, 0.5\}$. We observe that $\delta=0.07$ achieves the best few-shot segmentation performance. Nonetheless, we set $\delta$ to 0.1 to cover a wider range frequency band, which may bring in more robustness. 

\begin{table}[t]
\centering
\begin{adjustbox}{width=0.86\linewidth}
	\begin{tabular}{cccccc}
	\toprule
	\makecell*[c]{\textbf{Dimension}} & {30} & {60} & {90} & {120} & {144} \\
        \cmidrule(lr){1-1} \cmidrule(lr){2-6}
        \specialrule{0em}{1pt}{1pt}
		 $S_0$ & 26.33 & 48.76 & \textbf{51.42} & 51.13 & 51.20  \\
		 \specialrule{0em}{1pt}{1pt}
         $S_1$ & 31.65 & 50.24 & \textbf{54.94} & 54.87 & 53.19  \\
         $Avg$ & 28.99 & 49.50 & \textbf{53.18} & 53.00 & 52.19 \\ 
		 \specialrule{0em}{1pt}{1pt}
	\bottomrule
	\end{tabular}
\end{adjustbox}
\caption{\textbf{Ablation Study on Positional Encoding Dimension.} We report the results (\%) of TFS3D.}
\label{table:ablation_PEdim_TFS3D}
\end{table}

\begin{table}[t!]
\centering
\begin{adjustbox}{width=0.97\linewidth}
	\begin{tabular}{ccccccc}
	\toprule
	\makecell*[c]{\textbf{Dimension}} & {30} & {36} & {48} & {60} & {72} & {90} \\
        \cmidrule(lr){1-1} \cmidrule(lr){2-7}
        \specialrule{0em}{1pt}{1pt}
		 $S_0$ & 62.51 & 64.53 & 68.84 & \textbf{70.34} & 68.60 & 66.18 \\
		 \specialrule{0em}{1pt}{1pt}
         $S_1$ & 72.71 & 74.35 & \textbf{74.69} & 74.09 & 74.32 & 73.19 \\
         $Avg$ & 67.61 & 69.44 & 71.76 & \textbf{72.21} & 71.64 & 69.68 \\ 
		 \specialrule{0em}{1pt}{1pt}
	\bottomrule
	\end{tabular}
\end{adjustbox}
\caption{\textbf{Ablation Study on Positional Encoding Dimension.} We report the results (\%) of TFS3D-T.}
\label{table:ablation_PEdim_TFS3D-T}
\end{table}

\begin{table}[t!]
\centering
\begin{adjustbox}{width=0.97\linewidth}
	\begin{tabular}{ccccccc}
	\toprule
	\makecell*[c]{\textbf{$\theta$}} & {10} & {20} & {40} & {60} & {80} & {100}\\
    \cmidrule(lr){1-1} \cmidrule(lr){2-7}
    \specialrule{0em}{1pt}{1pt}
	TFS3D & 49.37 & \textbf{53.18} & 52.44 & 52.61 & 51.72 & 51.64 \\
	\specialrule{0em}{1pt}{1pt}
    TFS3D-T & 73.01 & 72.21 & \textbf{73.68} & 72.50 & 72.03 & 72.36\\
	\specialrule{0em}{1pt}{1pt}
	\bottomrule
	\end{tabular}
\end{adjustbox}
\caption{\textbf{Ablation for Parameter $\theta$ in Initial Layer} We report the average results (\%) of $S_0$ and $S_1$ split.}
\label{table:ablation_theta}
\end{table}

\begin{table}[t!]
\centering
\begin{adjustbox}{width=0.97\linewidth}
	\begin{tabular}{ccccccc}
	\toprule
	\makecell*[c]{\textbf{$\delta$}} & {0.01} & {0.05} & {0.07} & {0.1} & {0.2} & {0.5}\\
    \cmidrule(lr){1-1} \cmidrule(lr){2-7}
    \specialrule{0em}{1pt}{1pt}
	TFS3D & 45.46 & 48.30 & \textbf{53.34} & 53.18 & 50.66 & 36.29 \\
	\specialrule{0em}{1pt}{1pt}
    TFS3D-T & 62.57 & 67.03 & \textbf{73.43} & {72.21} & 73.06 & 70.01\\
	\specialrule{0em}{1pt}{1pt}
	\bottomrule
	\end{tabular}
\end{adjustbox}
\caption{\textbf{Ablation for Local Layer Parameter $\delta$.} We report the average results (\%) of $S_0$ and $S_1$ split.}
\label{table:ablation_delta}
\end{table}

\begin{figure}[t!]
\centering
\subfloat[Ablation for Kernel Size of Local Pooling in TFS3D-T]{\includegraphics[width=0.22\textwidth]{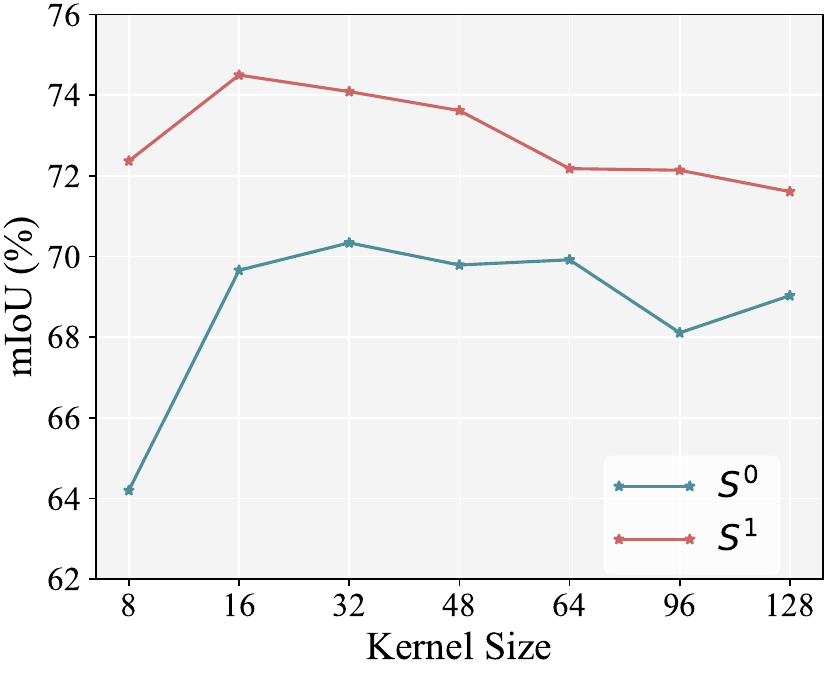}} 
\hspace{4pt}
\subfloat[Ablation of Linear Projection Numbers in TFS3D-T]{\includegraphics[width=0.22\textwidth]{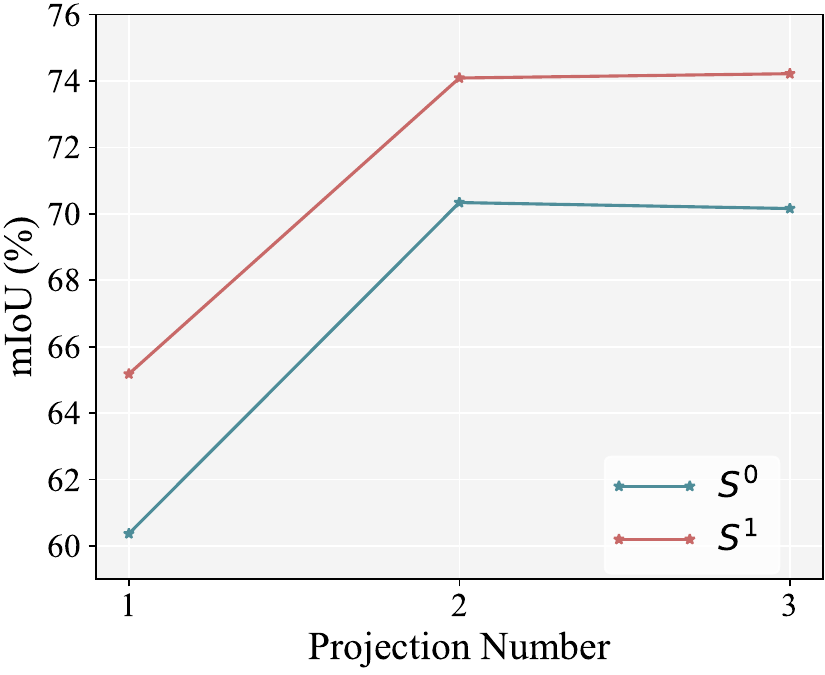}}
\vspace{-0.15cm}
\caption{\textbf{Ablation on Detailed Designs of TFS3D-T. 
}}
\label{fig:kernel_linear_QUEST}
\vspace{-0.3cm}
\end{figure}

\paragraph{Detailed Designs of TFS3D-T.} We mainly explore 2 hyperparameters in the QUEST module: the kernel size of the local maximum pooling and the number of linear projections in the fully connected layer. In Fig. \ref{fig:kernel_linear_QUEST}(a), we observe that the best performance is achieved when the kernel size is either 16 or 32. In Fig. \ref{fig:kernel_linear_QUEST}(b), experiments demonstrate that using 2 stacked linear projection operations is more beneficial for the final performance. A single linear projection is insufficient in capturing the necessary features for few-shot learning, while three or more projections may lead to overfitting.

\section{More Visualization}
In Fig. \ref{fig:eg_s3dis} and \ref{fig:eg_scannet}, we present more examples of 2-way-1-shot rooms from S3DIS and ScanNet datasets. It is worth noting that due to sparse sampling in certain regions of the rooms, some ScanNet rooms may appear incomplete, as shown in Fig. \ref{fig:eg_scannet}. All rooms are presented in a top-down view.

\begin{figure}[t!]
\centering
\includegraphics[width=0.47\textwidth]{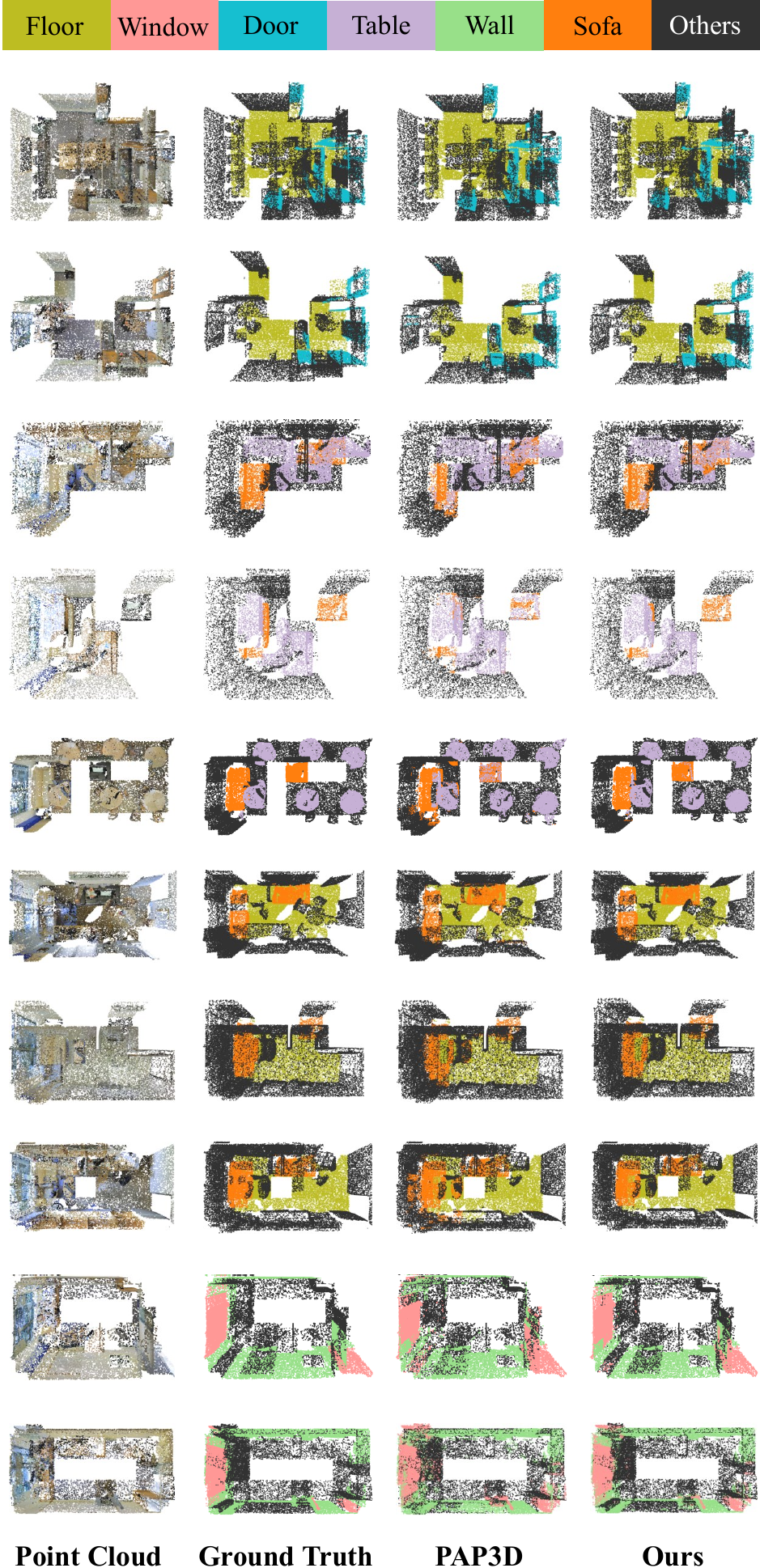}
\vspace{-0.3cm}
\caption{\textbf{Visualization of Results} on S3DIS dataset in 2-way-1-shot tasks. We present TFS3D-T's prediction in comparison to the ground truth and PAP3D model.}
\label{fig:eg_s3dis}
\vspace{-0.1cm}
\end{figure}

\begin{figure}[t]
\vspace{-0.0cm}
\centering
\includegraphics[width=0.473\textwidth]{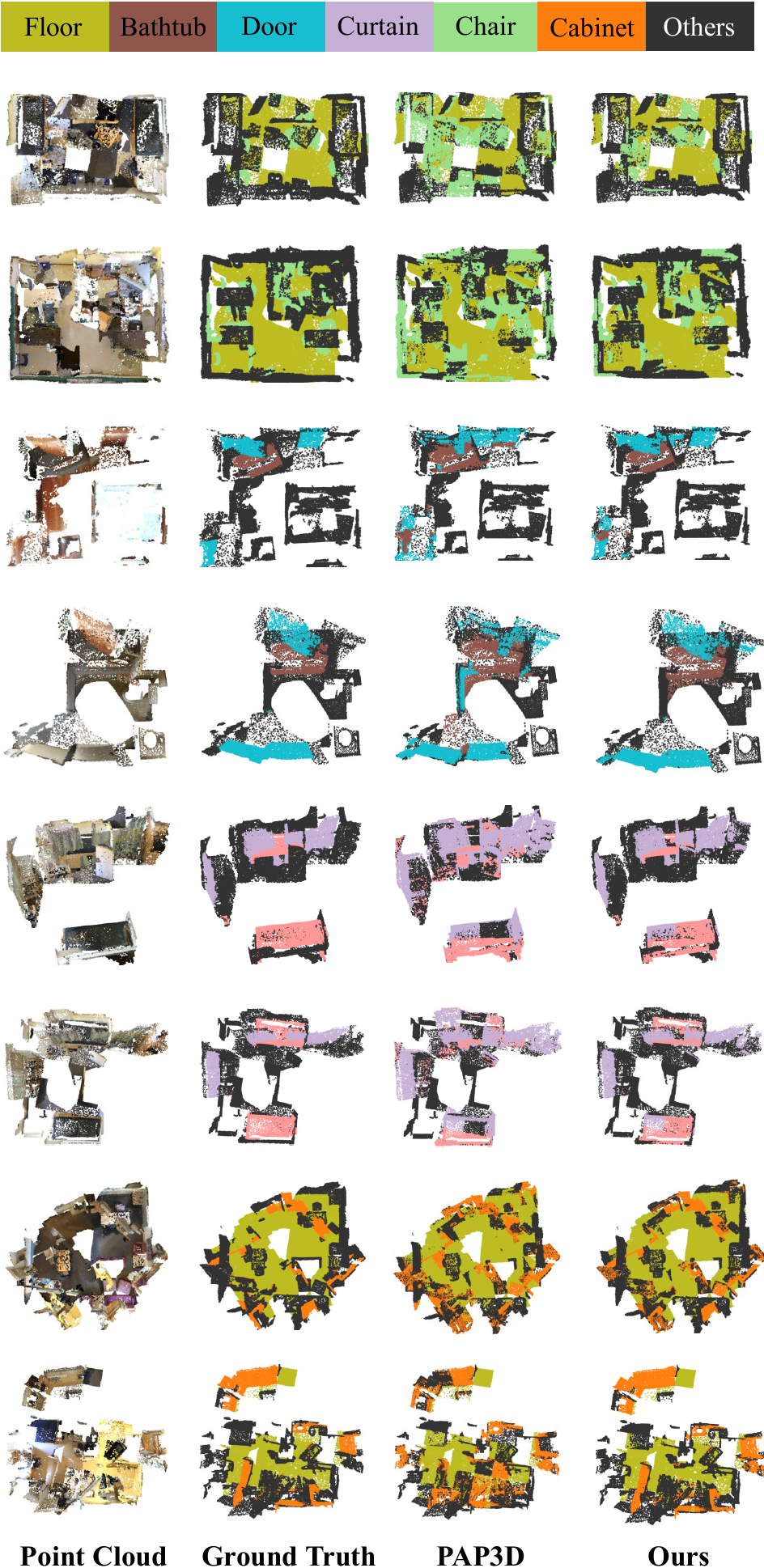}
\vspace{-0.3cm}
\caption{\textbf{Visualization of Results} on ScanNet dataset in 2-way-1-shot tasks. We present TFS3D-T's prediction in comparison to the ground truth and PAP3D model.}
\label{fig:eg_scannet}
\vspace{-0.1cm}
\end{figure}


\end{document}